\documentclass[letterpaper]{article}

\usepackage[preprint]{aaai2027}

\usepackage[hyphens]{url}
\usepackage{graphicx}
\usepackage{natbib}
\usepackage{caption}
\usepackage{booktabs}
\usepackage{multirow}
\usepackage{amsmath}
\usepackage{amssymb}
\usepackage{xspace}
\usepackage{hyperref}
\usepackage[capitalize,noabbrev]{cleveref}

\makeatletter
\DeclareRobustCommand\onedot{\futurelet\@let@token\@onedot}
\def\@onedot{\ifx\@let@token.\else.\null\fi\xspace}
\def\eg{\emph{e.g}\onedot} 
\makeatother

\newcommand{\OurMethod}{InstanceSplat\xspace}
\begin{document}

\title{\OurMethod: Instance-Aware Feed-Forward 3D Gaussian Splatting for Scene Understanding}

\author{
Minchao Jiang\textsuperscript{\rm 1,\rm 2},
Xiaoxuan Ma\textsuperscript{\rm 3},
Shunyu Jia\textsuperscript{\rm 4},
Haoru Wang\textsuperscript{\rm 5},\\
Zhang Liang\textsuperscript{\rm 4},
Wentao Zhu\textsuperscript{\rm 2}
}

\affiliations{
\textsuperscript{\rm 1}Shanghai Jiao Tong University,
\textsuperscript{\rm 2}Eastern Institute of Technology, Ningbo\\
\textsuperscript{\rm 3}Carnegie Mellon University,
\textsuperscript{\rm 4}Xidian University,
\textsuperscript{\rm 5}Peking University
}

\maketitle

\begin{abstract}
Feed-forward 3D Gaussian Splatting (3DGS) enables efficient and generalizable 3D reconstruction, but current feed-forward 3DGS methods for scene understanding remain largely category-oriented. 
In contrast, instance-aware 3DGS methods typically rely on per-scene optimization and often decouple reconstruction from instance and semantic learning, limiting reciprocal interactions among them. 
We present \OurMethod, a unified feed-forward 3DGS framework for generalizable 3D reconstruction and instance-aware scene understanding from pose-free multi-view images. 
In a single forward pass, \OurMethod constructs an instance-aware Gaussian representation that jointly encodes appearance, geometry, instance identity, and language-aligned semantics.
Shared 3D Gaussians ground instance identities across views, producing renderable and cross-view-consistent instance features. 
To allow reconstruction and scene understanding to benefit from each other, we further design an instance-centric learning strategy that connects reconstruction, instance learning, and semantic learning through shared instance structure.
Specifically, instance cues guide reconstruction, language-aligned semantics strengthen the discrimination of confusing same-category instances, and instance regions aggregate semantic evidence into coherent object-level predictions.
Experiments on novel-view synthesis, instance segmentation, and open-vocabulary semantic understanding under varying input-view settings and on an unseen dataset demonstrate state-of-the-art performance, practical efficiency, and strong generalization.  Project page: \href{https://jamchaos.github.io/InsSplat/}{InstanceSplat}.
\end{abstract}

\section{Introduction}
\label{sec:intro}
Structured understanding~\cite{yang2025thinking} of 3D environments supports a wide range of downstream applications, \eg robotic navigation~\cite{zheng20263dgsnav,anderson2018vision,hughes2024foundations,mao2025spatiallm} and physical interaction~\cite{stone2023open}. Among existing 3D representations~\cite{mildenhall2021nerf,foley1996computer}, 3D Gaussian Splatting (3DGS)~\cite{kerbl20233d} is particularly attractive due to its high-fidelity novel-view synthesis (NVS) and efficient rendering. However, standard 3DGS is designed primarily for visual reconstruction and does not explicitly encode instance structure or semantic information. This gap motivates representations that augment 3DGS with instance-level structure while preserving its rendering fidelity.
\begin{figure}[t]
    \centering
    \includegraphics[width=\columnwidth]{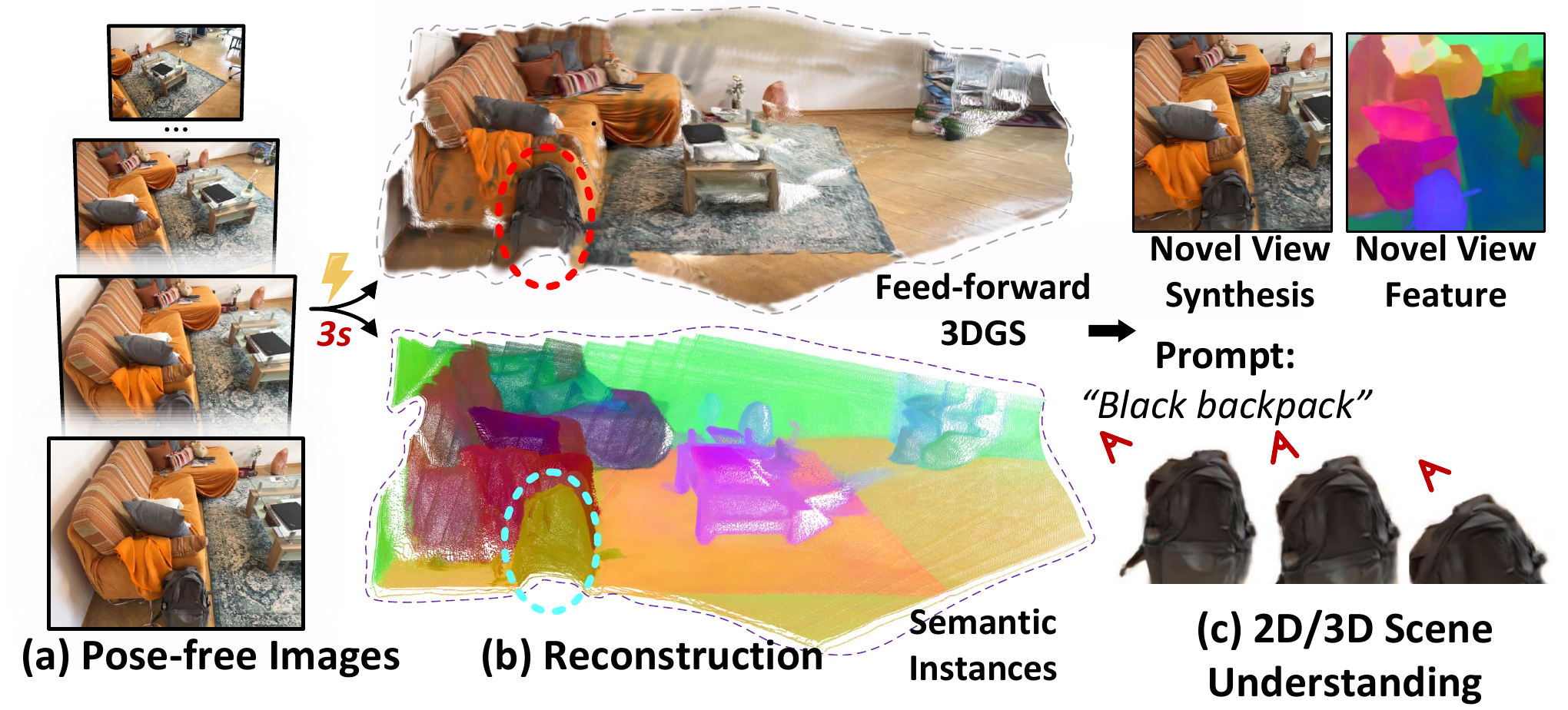}
    \caption{Given pose-free multi-view images (a), \OurMethod jointly predicts explicit 3D Gaussians with compact instance and open-vocabulary semantic features (b). Shared 3D geometry provides cross-view consistency, while instance cues enhance boundary-aware 3DGS reconstruction and semantic coherence. The resulting representation supports novel-view synthesis (NVS), instance segmentation, and open-vocabulary scene understanding (c).}
    \label{fig:teaser}
    \vspace{-6pt}
\end{figure}

Existing 3DGS methods for scene understanding mainly follow two paradigms. 
Scene-specific approaches augment an optimized Gaussian scene with semantic or instance features~\cite{qin2024langsplat,wu2024opengaussian,jiang2025votesplat,zhai2025panogs,li2025langsplatv2,li2025instancegaussian}. 
Although they enable semantic querying and instance segmentation, they rely on per-scene optimization and typically treat reconstruction and instance learning as separate stages, which limits efficiency and reciprocal interactions.
Feed-forward 3DGS methods instead reconstruct unseen scenes directly from images~\cite{charatan2024pixelsplat,chen2024mvsplat,tang2024hisplat,jiang2025anysplat,xu2025depthsplat}, while recent pose-free models further remove the requirement for calibrated cameras~\cite{ye2024no,jiang2025anysplat}. 
Recent scene-understanding extensions incorporate language-aligned features into these models~\cite{tian2025fleg,sun2025uni3r,fan2024large}, but their representations remain category-oriented and struggle to  distinguish object instances and preserve clear object boundaries.  
Consequently, existing methods offer either efficient and generalizable reconstruction without explicit instance modeling or instance-aware understanding through expensive per-scene optimization. A unified feed-forward framework that jointly learns reconstruction, instance structure, and semantics remains missing.

To this end, we propose \OurMethod, a unified feed-forward 3DGS framework for generalizable 3D reconstruction and instance-aware scene understanding from pose-free multi-view images. As illustrated in~\cref{fig:teaser}, \OurMethod maps the input views to an instance-aware Gaussian representation in a single forward pass. The representation jointly encodes appearance, geometry, instance identity, and language-aligned semantics, treating instance identity as an intrinsic Gaussian attribute rather than a feature learned after reconstruction. It therefore provides a unified and generalizable representation for NVS, instance segmentation, and open-vocabulary semantic understanding.

Furthermore, instance structure can serve as a shared interface between 3D reconstruction and semantic understanding.
To establish this interface, we introduce a 3D-Consistent Instance Grounding module that learns renderable and cross-view-consistent instance features. Compact instance embeddings attached to the shared Gaussians are differentiably rendered into supervised views, where prototype-based contrastive learning promotes intra-instance compactness, inter-instance separation, and cross-view alignment. Because observations from different views are coupled through the same 3D Gaussians, the resulting instance features are both renderable and consistent across views. Building on these grounded features, 
we introduce an Instance-Centric Coupling module, which enables reconstruction, instance learning, and semantic learning to reinforce one another. Instance cues guide boundary-aware 3DGS reconstruction, while instance structure and language-aligned semantics mutually refine each other: semantic cues improve the discrimination of same-category instances, and instance regions aggregate semantic evidence into coherent object-level predictions.

Our contributions are threefold:
\begin{enumerate}
\item We propose \OurMethod, a unified feed-forward 3DGS framework that constructs an instance-aware Gaussian representation from pose-free multi-view images, enabling instance-level scene understanding.

\item Based on this representation, we introduce 3D-Consistent Instance Grounding and Instance-Centric Coupling to establish cross-view-consistent instance structure and enable reciprocal interactions between 3D reconstruction and scene understanding.

\item Comprehensive experiments on novel-view synthesis, instance segmentation, and open-vocabulary semantic understanding under varying input-view settings and on an unseen dataset demonstrate SOTA performance, practical efficiency, and strong generalization.
\end{enumerate}

\section{Related Work}
\subsubsection{Scene-Specific 3DGS Scene Understanding.}
Early studies~\cite{zhi2021place,Siddiqui_2023_CVPR,kerr2023lerf} incorporate semantics into implicit neural fields for language-guided 3D understanding, while 3DGS~\cite{kerbl20233d} enables more efficient semantic feature fields~\cite{qin2024langsplat,li2025langsplatv2}. Instance-aware extensions provide finer-grained understanding: Gaussian Grouping~\cite{ye2024gaussian} attaches compact identity codes to Gaussians, Click-Gaussian~\cite{choi2024click} learns discriminative fields for interactive segmentation, and OpenGaussian~\cite{wu2024opengaussian} associates 3D-consistent instances with 2D CLIP features. InstanceGaussian~\cite{li2025instancegaussian} jointly models appearance and semantics for category-agnostic instance aggregation. Despite their different representations and supervision, these methods require per-scene optimization and often separate reconstruction, instance learning, and semantic assignment into stages, which limits efficiency and cross-scene generalization and prevents reciprocal interactions among these components.
\begin{figure*}[t]
    \centering
    \includegraphics[width=\textwidth]{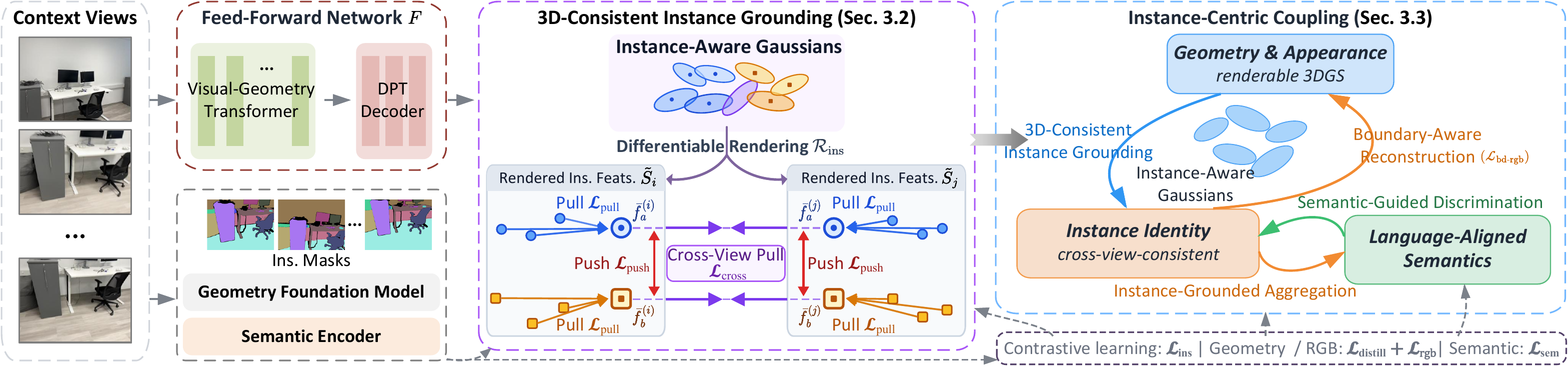}
    \caption{
    \OurMethod overview. Given pose-free multi-view images, a feed-forward backbone constructs an instance-aware Gaussian representation. By differentiable rendering instance features from shared Gaussians, 3D-Consistent Instance Grounding (\cref{sec:contrastive}) learns renderable and cross-view-consistent instance features.
    Instance-Centric Coupling (\cref{sec:mutual}) uses the grounded instance structure to guide boundary-aware reconstruction and enable reciprocal interactions between instance identity and language-aligned semantics.}
    \label{fig:pipeline}
    \vspace{-6pt}
\end{figure*}
\subsubsection{Feed-Forward 3DGS Scene Understanding.}
Feed-forward 3DGS methods directly predict Gaussian primitives from sparse images using multi-view matching, hierarchical modeling, and depth priors~\cite{Yu_2021_CVPR,Chen_2021_ICCV,chen2024mvsplat,tang2024hisplat,xu2025depthsplat}. Pose-free variants~\cite{ye2024no,smart2024splatt3r,jiang2025anysplat} remove the need for calibrated cameras, while query-based approaches~\cite{an2025c3g,ren2026tokengs} generate compact Gaussian sets with learnable tokens. Recent methods~\cite{fan2024large,sun2025uni3r,tian2025fleg} further introduce language-aligned semantic features for efficient open-vocabulary understanding, but remain primarily category-oriented and do not explicitly distinguish object instances. IGGT~\cite{li2025iggt} introduces instance-grounded contrastive learning into a feed-forward geometry transformer, but does not construct a renderable 3D Gaussian representation for joint novel-view synthesis and semantic understanding.
Concurrent work~\cite{yoo2026scenes} organizes Gaussians into object-centric token groups for reconstruction, segmentation, and manipulation. In contrast, \OurMethod attaches instance and semantic features to explicit Gaussians and couples reconstruction, instance learning, and semantic learning, enabling 3D-consistent instance grounding, boundary-aware reconstruction, and object-level semantic aggregation.

\section{Method}

\Cref{fig:pipeline} provides an overview of \OurMethod. Given pose-free multi-view images, our feed-forward framework first constructs an Instance-Aware Gaussian Representation (\cref{sec:architecture}). It then employs 3D-Consistent Instance Grounding (\cref{sec:contrastive}) to learn renderable and cross-view-consistent instance features. Building on the grounded instance structure, Instance-Centric Coupling (\cref{sec:mutual}) connects reconstruction, instance learning, and semantic understanding through reciprocal interactions.

\subsection{Instance-Aware Gaussian Representation}
\label{sec:architecture}
\paragraph{Problem Definition.}
Given $N$ pose-free input images $\{I_i\in\mathbb{R}^{H\times W\times 3}\}_{i=1}^{N}$, our objective is to construct an instance-aware 3DGS representation with open-vocabulary semantics.
The feed-forward network $F$ jointly predicts camera parameters $t_i$, including camera intrinsics and extrinsics, depth maps $D_i\in\mathbb{R}^{H\times W}$, point maps $P_i\in\mathbb{R}^{H\times W\times3}$, and a dense Gaussian set $\mathcal{G}$:
\begin{equation}
F : \{I_i\}_{i=1}^N \rightarrow \{(t_i,D_i,P_i)\}_{i=1}^N,\mathcal{G}.
\end{equation}

\subsubsection{Pose-Free Gaussian Reconstruction Backbone.}
Following VGGT~\cite{wang2025vggt}, we use a pretrained DINOv2 backbone~\cite{oquab2023dinov2} and a 24-layer Transformer with alternating within-view and cross-view attention to encode the input views. A DPT-style geometry decoder~\cite{Ranftl_2021_ICCV} predicts $t_i$, $P_i$, and $D_i$, from which Gaussian centers are recovered by back-projection. A second DPT-style decoder regresses the remaining Gaussian attributes together with compact semantic features. These components provide the pose-free reconstruction foundation on which we construct the instance-aware representation. Following IGGT~\cite{li2025iggt}, an instance decoder predicts a pixel-aligned feature map $S_i\in\mathbb{R}^{H\times W\times8}$ from the multi-level Transformer features and incorporates local geometric cues through sliding-window cross-attention. The resulting embeddings are attached to the Gaussian primitives derived from the corresponding pixels.

\subsubsection{Instance-Aware Gaussian Construction.}
For each input pixel, we combine the recovered center, Gaussian attributes, instance embedding, and compact semantic feature into a single primitive. The resulting dense representation is
\begin{equation}
\mathcal{G} = \{(\mu_k, R_k, s_k, c_k, \alpha_k, C_k, S_k, L_k)\}_{k=1}^{N\times H\times W},
\end{equation}
where $k$ indexes the per-pixel Gaussians across all input views. Specifically, $\mu_k \in \mathbb{R}^3$ denotes the 3DGS position. $R_k\in \mathbb{R}^{3 \times 3}$ represents the rotation matrix, $s_k \in \mathbb{R}^{3}$ denotes the scale, $c_k\in \mathbb{R}^{3}$ denotes the color, $\alpha_k \in [0, 1]$ denotes the opacity, and $C_k \in \mathbb{R}$ is a learned confidence score.
Assigning one Gaussian per pixel inevitably leads to severe spatial redundancy and computational overhead when scaling to dense multi-view settings. To mitigate this, following AnySplat~\cite{jiang2025anysplat}, we employ a confidence-aware voxelization strategy to merge primitives and obtain a voxel-aligned 3DGS set $\mathcal{G}'$.
Each primitive additionally carries an instance embedding $S_k\in\mathbb{R}^{8}$ and a semantic feature $L_k\in\mathbb{R}^{d_s}$. The confidence-aware aggregation also fuses these features, yielding a compact representation that preserves appearance, geometry, instance identity, and language-aligned semantics in a shared 3D carrier.

\subsection{3D-Consistent Instance Grounding}
\label{sec:contrastive}
Given the instance-aware Gaussians, we learn instance identity through rendered-space supervision. Unlike contrastive learning on independent 2D feature maps, our supervision is mediated by shared Gaussians, grounding image-space instance labels in an explicit and renderable 3D representation.

\subsubsection{Differentiable Rendering.}
We render the instance embeddings attached to the voxelized Gaussians into each supervised view using the same alpha-compositing process as RGB rendering:
\begin{equation}
\tilde{S}_i = \mathcal{R}_{\text{ins}}(\mathcal{G}', t_i),
\end{equation}
where $\tilde{S}_i\in\mathbb{R}^{H\times W\times8}$ is the rendered instance feature map and $\mathcal{R}_{\text{ins}}$ denotes differentiable instance feature rendering. Observations from different viewpoints are therefore coupled through the same 3D primitives, allowing image-space supervision to establish consistent instance identities in 3D.

\subsubsection{Rendered-Space Contrastive Supervision.}
Let $M_i$ denote the GT instance mask with instance IDs aligned across views, and let $\mathcal{K}_i$ denote the set of valid instance IDs visible in that view. We denote $\ell_2$ normalization by $\operatorname{norm}(x)=x/\|x\|_2$. We apply it to the rendered feature at each valid pixel $p$ and average the normalized features within each visible instance to obtain its prototype:
\begin{equation}
f_i(p)=\operatorname{norm}(\tilde{S}_i(p)),
\quad
\bar{f}_k^{(i)}=
\frac{1}{|\Omega_k^{(i)}|}
\sum_{p\in\Omega_k^{(i)}}f_i(p),
\end{equation}
where $\Omega_k^{(i)}$ is the set of valid pixels belonging to instance $k$ in view $i$. 

As illustrated in~\cref{fig:pipeline}, we use a prototype-based objective to promote intra-instance compactness, inter-instance separation, and cross-view consistency. Here, $[x]_+=\max(0,x)$ and $\delta_{\text{pull}}$, $\delta_{\text{push}}$, and $\delta_{\text{cross}}$ are margin hyperparameters. Within each view, the pull term keeps pixels close to their instance prototype:
\begin{equation}
\label{eq:pull}
\mathcal{L}_{\text{pull}}^i =
\frac{1}{\lvert \mathcal{K}_i \rvert}\sum_{k\in\mathcal{K}_i}
\mathbb{E}_{p \in \Omega_k^{(i)}}
\left[\|f_i(p)-\bar{f}_k^{(i)}\|_2-\delta_{\text{pull}}\right]_+.
\end{equation}

The push term separates different instance prototypes:
\begin{equation}
\label{eq:push}
\mathcal{L}_{\text{push}}^i =
\mathbb{E}_{\substack{k,l\in\mathcal{K}_i\\k<l}}
\eta_{k,l}^{(i)}
\left[
\delta_{\text{push}}
-\|\bar{f}_k^{(i)}-\bar{f}_l^{(i)}\|_2
\right]_+.
\end{equation}
Here, $\eta_{k,l}^{(i)}=1$ for uniform separation and is instantiated by semantic similarity in~\cref{sec:semantic_discrimination} for the full model.

For instances observed in multiple views, we use a cross-view term to align their prototypes:
\begin{equation}
\label{eq:cross}
\mathcal{L}_{\mathrm{cross}}
=
\mathbb{E}_{\substack{
i<j\\
k\in\mathcal{K}_i\cap\mathcal{K}_j
}}
\left[
\|\bar{f}_k^{(i)}-\bar{f}_k^{(j)}\|_2-\delta_{\mathrm{cross}}
\right]_+.
\end{equation}

The global terms $\mathcal{L}_{\text{pull}}$ and $\mathcal{L}_{\text{push}}$ denote the averages of $\mathcal{L}_{\text{pull}}^i$ and $\mathcal{L}_{\text{push}}^i$ over all supervised views. The complete instance grounding objective is
\begin{equation}
\mathcal{L}_{\text{ins}}=
\lambda_{\text{pull}} \mathcal{L}_{\text{pull}}
+ \lambda_{\text{push}} \mathcal{L}_{\text{push}}
+ \lambda_{\text{cross}} \mathcal{L}_{\text{cross}}.
\end{equation}
The resulting embeddings remain consistent across input views and can be rendered from novel viewpoints. We next use this grounded instance structure to couple reconstruction with semantic understanding.

\begin{figure}[t]
    \centering
    \includegraphics[width=\columnwidth]{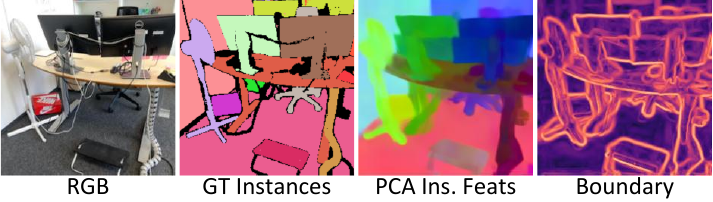}
    \caption{Visualization of instance-derived boundary weights.  Brighter responses indicate larger weights near instance boundaries. }
    \label{fig:boundary}
    \vspace{-6pt}
\end{figure}
\begin{table*}[t]
\centering
\resizebox{\textwidth}{!}{
\begin{tabular}{l|ccccc|ccccc}
\toprule
\multirow{2}{*}{Methods}
& \multicolumn{5}{c|}{2 views}
& \multicolumn{5}{c}{4 views} \\
\cmidrule(lr){2-6} \cmidrule(lr){7-11}
& mIoU$\uparrow$ & mAcc$\uparrow$ & PSNR$\uparrow$ & SSIM$\uparrow$ & LPIPS$\downarrow$
& mIoU$\uparrow$ & mAcc$\uparrow$ & PSNR$\uparrow$ & SSIM$\uparrow$ & LPIPS$\downarrow$ \\
\midrule
AnySplat~\cite{jiang2025anysplat}
& -- & -- & \underline{20.76} & \underline{0.7354} & \underline{0.3394}
& -- & -- & \underline{20.73} & 0.7150 & \underline{0.3256} \\
LSeg~\cite{li2022languagedriven}
& \underline{48.16} & \underline{81.62} & -- & -- & --
& \underline{37.00} & \textbf{75.75} & -- & -- & -- \\
LSM~\cite{fan2024large}
& 22.84 & 63.55 & 12.61 & 0.5686 & 0.5461
& -- & -- & -- & -- & -- \\
Uni3R~\cite{sun2025uni3r}
& 27.36 & 67.36 & 15.59 & 0.6499 & 0.4483
& -- & -- & -- & -- & -- \\
C3G~\cite{an2025c3g}
& 46.14 & 76.12 & 19.21 & 0.7168 & 0.4369
& 33.15 & 68.43 & 20.69 & \underline{0.7180} & 0.4051 \\
\midrule
Ours w/o Hard Neg.
& 65.01 & 85.03 & 20.99 & 0.7446 & 0.3338
& \textbf{45.96} & \underline{73.77} & 21.44 & 0.7343 & 0.3251 \\
Ours w/o Sem. Agg.
& 59.91 & 81.03
& \multirow{2}{*}{\textbf{21.20}}
& \multirow{2}{*}{\textbf{0.7497}}
& \multirow{2}{*}{\textbf{0.3266}}
& 42.01 & 72.87
& \multirow{2}{*}{\textbf{21.62}}
& \multirow{2}{*}{\textbf{0.7370}}
& \multirow{2}{*}{\textbf{0.3200}} \\
\textbf{Ours}
& \textbf{65.23} & \textbf{85.22} & & &
& 43.51 & 73.68 & & & \\
\midrule
\multirow{2}{*}{Methods}
& \multicolumn{5}{c|}{8 views}
& \multicolumn{5}{c}{16 views} \\
\cmidrule(lr){2-6} \cmidrule(lr){7-11}
& mIoU$\uparrow$ & mAcc$\uparrow$ & PSNR$\uparrow$ & SSIM$\uparrow$ & LPIPS$\downarrow$
& mIoU$\uparrow$ & mAcc$\uparrow$ & PSNR$\uparrow$ & SSIM$\uparrow$ & LPIPS$\downarrow$ \\
\midrule
AnySplat~\cite{jiang2025anysplat}
& -- & -- & 21.14 & 0.7402 & \underline{0.3289}
& -- & -- & \underline{21.53} & \underline{0.7589} & \underline{0.3358} \\
LSeg~\cite{li2022languagedriven}
& 42.53 & \underline{78.64} & -- & -- & --
& \underline{45.29} & \underline{80.51} & -- & -- & -- \\
Uni3R~\cite{sun2025uni3r}
& 33.28 & 70.53 & 18.19 & 0.7237 & 0.3871
& 33.37 & 71.91 & 18.01 & 0.7381 & 0.4025 \\
C3G~\cite{an2025c3g}
& \underline{44.68} & 75.25 & \underline{21.70} & \textbf{0.7642} & 0.3910
& 40.82 & 73.51 & 18.86 & 0.7257 & 0.4685 \\
\midrule
Ours w/o Hard Neg.
& 58.33 & 79.88 & 21.54 & 0.7520 & 0.3281
& 64.33 & 84.11 & 21.88 & 0.7695 & 0.3365 \\
Ours w/o Sem. Agg.
& 52.85 & 74.72
& \multirow{2}{*}{\textbf{21.82}}
& \multirow{2}{*}{\underline{0.7593}}
& \multirow{2}{*}{\textbf{0.3188}}
& 57.70 & 77.73
& \multirow{2}{*}{\textbf{22.13}}
& \multirow{2}{*}{\textbf{0.7765}}
& \multirow{2}{*}{\textbf{0.3288}} \\
\textbf{Ours}
& \textbf{58.76} & \textbf{80.08} & & &
& \textbf{64.60} & \textbf{84.29} & & & \\
\bottomrule
\end{tabular}
}
\caption{Quantitative comparison with feed-forward 3DGS SOTA methods and ablations on ScanNet under varying input views. }
\label{tab:multi_view_results}
\vspace{-6pt}
\end{table*}
\subsection{Instance-Centric Coupling}
\label{sec:mutual}
As illustrated in~\cref{fig:pipeline}, once grounded in the explicit Gaussians, instance structure serves as a shared interface between reconstruction and semantic understanding. 
Local instance discontinuities guide reconstruction toward object boundaries, language-aligned semantic cues emphasize confusing same-category objects during instance learning, and instance regions provide stable object-level units for semantic aggregation.

\subsubsection{Boundary-Aware Reconstruction.}
Feed-forward reconstruction models commonly use a DPT-style head with convolutional feature fusion and bilinear upsampling to predict dense geometry. Its spatial smoothness is beneficial within a continuous surface, but can blur sharp depth changes between adjacent instances. The resulting geometry and Gaussian placement are therefore less reliable near object boundaries, where small spatial errors can substantially degrade RGB compositing. To focus reconstruction on these challenging regions, we derive a continuous boundary score from the pixel-aligned instance feature map $S_i$ before Gaussian rasterization.

For each pixel $p$, we normalize its instance feature as $z_i(p)=\operatorname{norm}(S_i(p))$ and compute the largest cosine distance to its four-connected neighbors:
\begin{equation}
r_i(p)=\max_{q\in\mathcal{N}_1(p)}
\left(1-z_i(p)^\top z_i(q)\right),
\end{equation}
where $\mathcal{N}_1(p)$ is the local neighborhood of $p$. Features within an instance have a small cosine distance, whereas a large value indicates a likely instance boundary.
We expand this response locally and map it to a soft boundary weight:
\begin{equation}
b_i(p)=
\sigma\!\left(
\frac{\max_{q\in\mathcal{N}_w(p)}r_i(q)-\tau}{T}
\right),
\end{equation}
where $w=2$ corresponds to a $5\times5$ window, $\tau=0.15$, and $T=0.05$. As shown in~\cref{fig:boundary}, the resulting weights closely follow instance boundaries while remaining low within coherent instance regions.
Let $\rho_i(p)$ denote the channel-averaged Charbonnier discrepancy~\cite{charbonnier1994two} between the rendered color $\hat{I}_i(p)$ and the GT color $I_i(p)$, with $\epsilon=10^{-3}$.
The instance-boundary-aware RGB loss is
\begin{equation}
\mathcal{L}_{\text{bd-rgb}}=
\frac{1}{N}\sum_{i=1}^{N}
\frac{\sum_p b_i(p)\rho_i(p)}
{\sum_p b_i(p)}.
\label{eq:bd-rgb}
\end{equation}
It relies only on learned instance features and thus requires neither GT instance boundaries nor opacity-based gating.

\subsubsection{Semantic-Guided Instance Discrimination.}
\label{sec:semantic_discrimination}
We render compact semantic features and project them to the language-aligned teacher space:
\begin{equation}
\tilde{L}_i=\mathcal{R}_{\text{sem}}(\mathcal{G}',t_i),
\qquad E_i(p)=\Phi_{\text{sem}}(\tilde{L}_i(p)),
\end{equation}
where $\mathcal{R}_{\text{sem}}$ denotes semantic feature rendering and $\Phi_{\text{sem}}$ is a learned projection. Given a teacher feature map  $Y_i$ produced by the Semantic Encoder~\cite{li2022languagedriven}, semantic alignment is enforced by
\begin{equation}
\mathcal{L}_{\text{sem}}=
\mathbb{E}_{i,p}\left[
1-\operatorname{norm}(E_i(p))^\top
\operatorname{norm}(Y_i(p))
\right].
\end{equation}
We construct an instance-level semantic prototype by pooling the aligned features within each GT instance:
\begin{equation}
\bar{\ell}_k^{(i)}
=
\operatorname{norm}\!\left(
\mathbb{E}_{p\in\Omega_k^{(i)}}
\left[\operatorname{norm}(E_i(p))\right]
\right).
\end{equation}

The semantic similarity between two instances defines the weight in~\cref{eq:push}:
\begin{equation}
\label{eq:semantic_weight}
\eta_{k,l}^{(i)}
=1+(\eta_{\max}-1)
\left[
\frac{(\bar{\ell}_k^{(i)})^\top\bar{\ell}_l^{(i)}-\tau_{\text{sem}}}
{1-\tau_{\text{sem}}}
\right]_0^1,
\end{equation}
where $[x]_0^1=\min(1,\max(0,x))$. Semantically dissimilar pairs retain unit weight, whereas semantically similar instance pairs receive larger separation weights. Instance and semantic source features are detached in their respective coupling paths.

\subsubsection{Instance-Grounded Semantic Aggregation.}
At inference, we render the instance and semantic feature maps from a target viewpoint and cluster the normalized instance embeddings using HDBSCAN~\cite{mcinnes2017hdbscan}, obtaining object regions $\{\hat{\Omega}_k^{(i)}\}$. Rather than assigning semantic labels independently at each pixel, we aggregate the aligned semantic features within each predicted instance:
\begin{equation}
\hat{\ell}_k^{(i)}
=
\operatorname{norm}\!\left(
\mathbb{E}_{p\in\hat{\Omega}_k^{(i)}}
\left[\operatorname{norm}(E_i(p))\right]
\right).
\end{equation}
The resulting prototype provides a stable object-level semantic representation. We assign an open-vocabulary label by comparing $\hat{\ell}_k^{(i)}$ with text embeddings and propagate the selected label to the entire instance region. This instance-grounded aggregation suppresses local semantic noise and produces coherent predictions within each object while preserving boundaries between different instances.

\paragraph{Training Objective.}
The overall objective is defined as
\begin{equation}
\mathcal{L}=\mathcal{L}_{\text{rgb}}+\mathcal{L}_{\text{distill}}
+\lambda_{\text{ins}}\mathcal{L}_{\text{ins}}
+\lambda_{\text{sem}}\mathcal{L}_{\text{sem}}
+\lambda_{\text{bd}}\mathcal{L}_{\text{bd-rgb}},
\end{equation}
where $\mathcal{L}_{\text{rgb}}$ denotes the photometric reconstruction loss $\mathcal{L}_{\text{rgb}} =
\text{MSE}(I_i, \hat{I}_i)+ \lambda_p \, \text{Perceptual}(I_i, \hat{I}_i)$, and $\mathcal{L}_{\text{distill}}$ enforces consistency between the predicted geometry, including camera poses and depth maps, and the pseudo-supervision provided by the Geometry Foundation Model~\cite{wang2025vggt}. Additional details are provided in Section~C of the supplementary material.

\begin{figure*}[t]
    \centering
    \includegraphics[width=\textwidth]{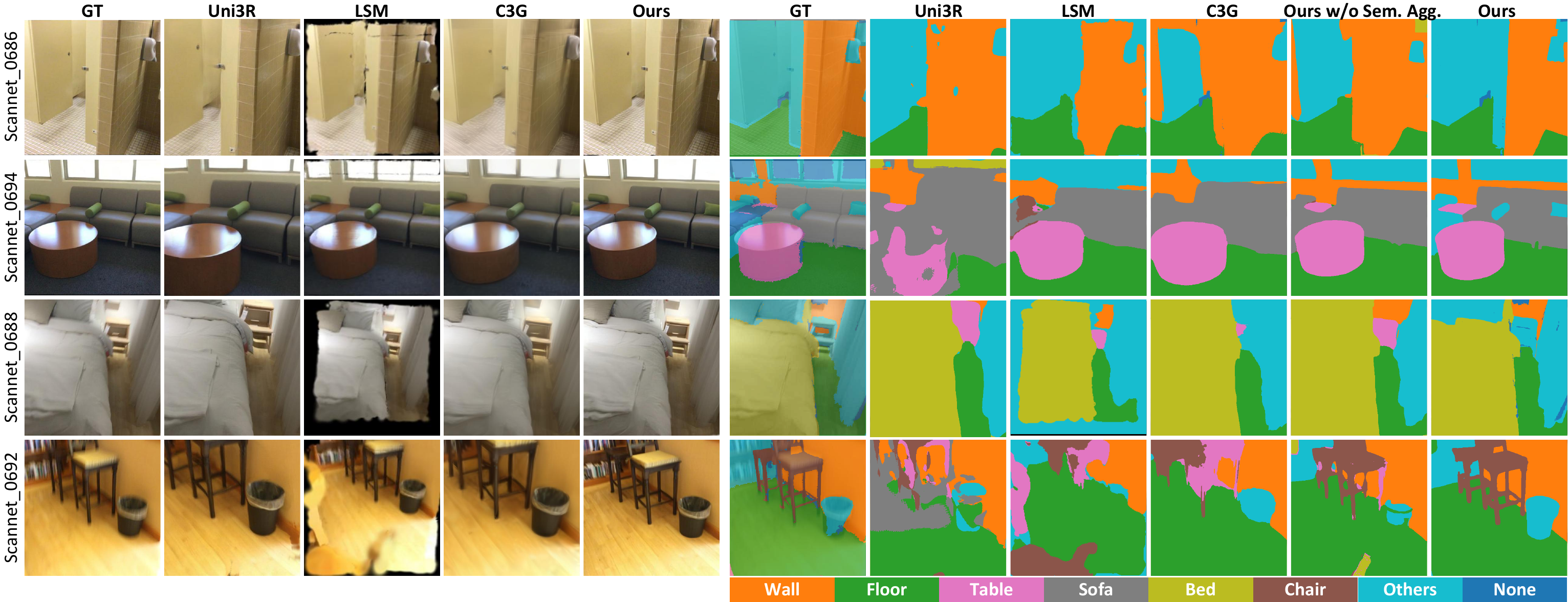}
    \caption{Qualitative comparison with feed-forward 3DGS baselines on novel-view rendering and semantic segmentation. We additionally show \OurMethod without instance-level semantic aggregation.}
    \label{fig:uni3re1}
    \vspace{-6pt}
\end{figure*}
\begin{figure}[t]
\centering
\includegraphics[width=\linewidth]{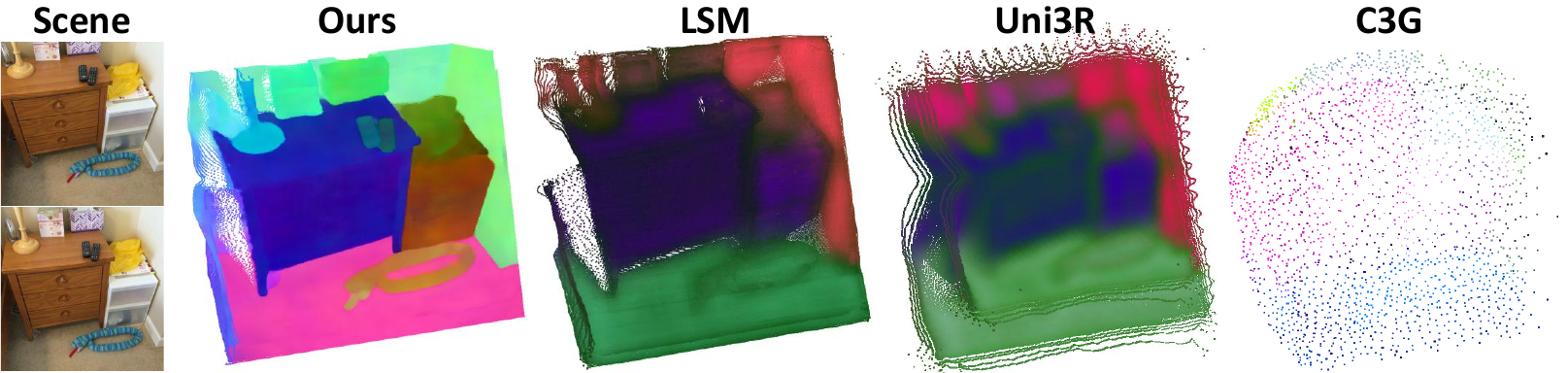}
\caption{Visualization of the 3D feature fields after PCA projection.
For \OurMethod, we visualize instance features, whereas for the other methods, we visualize semantic features.}
\label{fig:PCA}
\vspace{-6pt}
\end{figure}
\section{Experiment}
\subsection{Experimental Setup}
\subsubsection{Evaluation Details.}
We train our model on a mixture of ScanNet++~\cite{yeshwanth2023scannet++} and ScanNet~\cite{dai2017scannet} (1,565 scenes in total).
For ScanNet++, we use the InsScene-15K variant released by IGGT~\cite{li2025iggt}, where instance masks are further refined for higher-quality supervision.
Following LSM~\cite{fan2024large} and Uni3R~\cite{sun2025uni3r}, we evaluate on 50 held-out ScanNet scenes and compute PSNR, SSIM, and LPIPS on rendered novel views, together with mIoU and mAcc for novel-view segmentation.
To align with IGGT's training protocol, we additionally incorporate the RE10K subset released in InsScene-15K (5,137 scenes) and evaluate temporal instance segmentation consistency on the benchmark provided by IGGT using Temporal mIoU (T-mIoU) and Temporal Success Rate (T-SR), which quantify the cross-view consistency induced by the 3DGS representation.

\subsubsection{Implementation Details.}
During training, input images are resized such that the longer side is at most 448 pixels; the aspect ratio is randomly sampled in $[0.5, 1.0]$.
For each iteration, we sample 2--8 context views per scene.
Training is performed on $2\times$A800 GPUs with a batch size of 8 images.
We use AdamW with a base learning rate of $1\times10^{-4}$; the learning rate of Transformer layers is scaled by 0.1.
We apply 1k warm-up iterations followed by a cosine decay schedule for the remaining 20k iterations. Training completes in approximately 21 hours. 
We set $\lambda_p=0.05$, $\lambda_{\mathrm{ins}}=0.01$, $\lambda_{\mathrm{sem}}=\lambda_{\mathrm{bd}}=0.02$, and $(\lambda_{\mathrm{pull}},\lambda_{\mathrm{push}},\lambda_{\mathrm{cross}})=(2,1,2)$.

\subsection{Comparison to SOTA Methods}
\subsubsection{Comparison to Feed-forward Methods.}
Following the evaluation protocol of Uni3R~\cite{sun2025uni3r}, \cref{tab:multi_view_results} compares all methods under 2-, 4-, 8-, and 16-view settings. For a fair comparison, all metrics are computed at a resolution of $448\times448$. Among methods reporting reconstruction metrics, \OurMethod achieves the best PSNR and LPIPS. Among non-ablation methods, \OurMethod consistently obtains the highest mIoU across all settings, demonstrating robustness to the number of context views. The slightly lower mAcc in the 4-view setting is partly attributable to imbalanced class support induced by target-view sampling. In particular, the ceiling class is rarely observed, causing greater variation in the macro-averaged metric.
As shown in~\cref{fig:uni3re1}, \OurMethod produces cleaner novel-view semantic masks whose boundaries align more closely with object contours. 
\cref{fig:PCA} compares the learned 3DGS feature fields and reconstructed geometry. After PCA projection, the instance features of \OurMethod form coherent yet distinct representations for individual objects, whereas category-level semantic features tend to merge different instances of the same class, as illustrated by the two drawers in the scene. The reconstructed Gaussian distributions also reveal clear geometric differences. Uni3R exhibits a layered structure indicative of inaccurate depth estimation, while C3G suffers from severe geometric distortions and fails to preserve the scene structure. In contrast, \OurMethod produces a geometrically coherent reconstruction with well-separated, instance-aware features.

\begin{table}[t]
\centering
\resizebox{\columnwidth}{!}{
\begin{tabular}{l|ccc|ccc}
\toprule
\multirow{2}{*}{Methods}
& \multicolumn{3}{c|}{8 views}
& \multicolumn{3}{c}{16 views} \\
\cmidrule(lr){2-4}\cmidrule(lr){5-7}
& Time$\downarrow$ & mIoU$\uparrow$ & Acc@0.25$\uparrow$
& Time$\downarrow$ & mIoU$\uparrow$ & Acc@0.25$\uparrow$ \\
\midrule
InsGa
& 254~min & 10.72 & 11.76
& 280~min & 38.72 & 47.06 \\
OpenGa
& 55~min & 26.23 & 35.29
& 61~min & 51.11 & 64.71 \\
\textbf{Ours}
& \textbf{2.75~s} & \textbf{63.22} & \textbf{80.00}
& \textbf{3.65~s} & \textbf{53.85} & \textbf{73.33} \\
\bottomrule
\end{tabular}
}
\caption{Quantitative comparison with OpenGaussian (OpenGa)~\cite{wu2024opengaussian} and InstanceGaussian (InsGa)~\cite{li2025instancegaussian} under different numbers of context views on the LERF~\cite{kerr2023lerf} dataset (unseen during training). Following OpenGaussian, we report mean mask IoU (mIoU) and the fraction of instances with IoU above 0.25 (Acc@0.25).
}
\label{tab:OpenGaussian}
\vspace{-6pt}
\end{table}
\begin{figure*}[t]
    \centering
    \includegraphics[width=\linewidth]{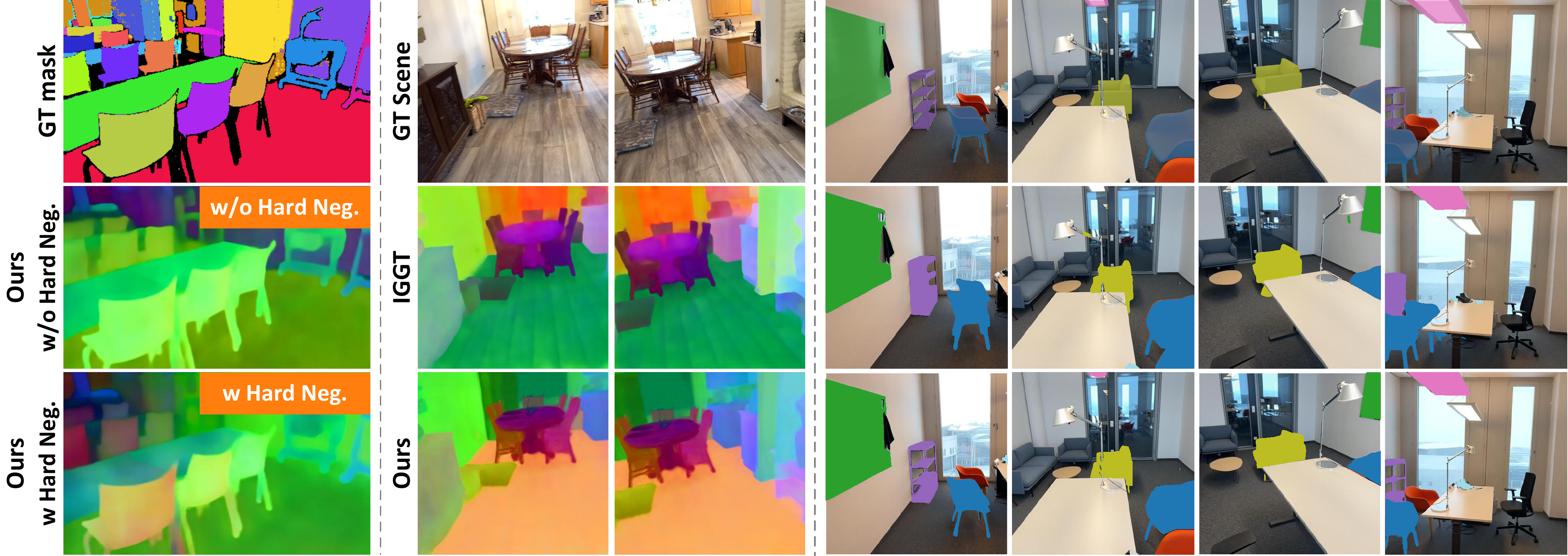}
    \caption{Qualitative analysis of instance representations. The left panel shows the effect of semantic-aware hard-negative reweighting, with the GT mask for reference. The middle and right panels compare PCA-projected features and track masks between IGGT and \OurMethod, respectively. }
    \label{fig:IGGT_vis}
    \vspace{-2pt}
\end{figure*}
\subsubsection{Comparison to Per-Scene Optimized Methods.}
To evaluate both efficiency and generalization, we additionally compare against per-scene optimized methods.
These approaches typically rely on Structure-from-Motion (SfM)~\cite{Schonberger_2016_CVPR} to estimate camera poses and adopt a staged pipeline that reconstructs the scene first and then learns instance features, resulting in high computational overhead and limited scalability.
For each view with a GT instance mask in LERF, we reserve the annotated view for evaluation and select 8 or 16 nearby frames as input views. Per-scene baselines optimize their Gaussians on the selected inputs, whereas \OurMethod reconstructs 3DGS and instance features from the same pose-free inputs in a single forward pass, reducing the total reconstruction and feature assignment time to roughly 3 seconds.
Although LERF is unseen during training, ours outperforms all baselines under both settings, demonstrating strong cross-dataset generalization. Its performance decreases from 8 to 16 input views, likely because the domain shift reduces pose accuracy and causes cross-view misalignment to accumulate as additional views are introduced.

\begin{table}[t]
\centering
\resizebox{\columnwidth}{!}{
\begin{tabular}{lccccc}
\toprule
Methods & T-mIoU$\uparrow$ & T-SR$\uparrow$ & PSNR$\uparrow$ & SSIM$\uparrow$ & LPIPS$\downarrow$ \\
\midrule
IGGT~\cite{li2025iggt} & 61.99 & 45.15 & -- & -- & -- \\
Ours (3DGS frozen) & 58.86 & 48.10 & 21.66 & 0.775 & 0.293 \\
Ours w/o $\mathcal{L}_{\mathrm{cross}}$ & 54.23 & 31.46 & 21.69 & \textbf{0.813} & 0.309 \\
\textbf{Ours} & \textbf{64.03} & \textbf{51.04} & \textbf{22.19} & 0.781 & \textbf{0.286} \\
\bottomrule
\end{tabular}}
\caption{Comparison with IGGT and ablations of 3DGS joint optimization and the cross-view consistency term in~\cref{eq:cross} on the benchmark provided by IGGT. T-SR@0.5.}
\label{tab:IGGT}
\vspace{-6pt}
\end{table}

\subsection{Comparison to IGGT}
Since IGGT~\cite{li2025iggt} does not support NVS, we compare it with \OurMethod on the context views. The higher T-mIoU and T-SR in~\cref{tab:IGGT} indicate that grounding instance features in a shared and renderable 3DGS representation strengthens the cross-view consistency of instance tracks.
The qualitative results in~\cref{fig:IGGT_vis} also support this interpretation. 
Compared with IGGT, \OurMethod preserves thin structures such as chair legs and produces smoother, more coherent instance features over the floor region. Our tracked instance masks also retain more accurate object structures across views, particularly for the cabinet and the chair.

\begin{table}[t]
\centering
\resizebox{\columnwidth}{!}{
\begin{tabular}{lccccc}
\toprule
Methods & RMSE$\downarrow$ & AbsRel$\downarrow$ & PSNR$\uparrow$ & SSIM$\uparrow$ & LPIPS$\downarrow$ \\
\midrule
Ours w/o $\mathcal{L}_{\text{bd-rgb}}$ & {0.2375} & 0.0995 & 22.11 & 0.7736 & \textbf{0.3288} \\
\textbf{Ours} & \textbf{0.1985} & \textbf{0.0911} & \textbf{22.13} & \textbf{0.7765} & \textbf{0.3288} \\
\bottomrule
\end{tabular}}
\caption{Ablation of the instance-boundary-aware RGB objective on the Uni3R benchmark.}
\label{tab:Ablation}
\vspace{-6pt}
\end{table}
\subsection{Ablation Study}
\paragraph{Gaussians for 3D-Consistent Instance Grounding.}
\Cref{tab:IGGT} evaluates how the Gaussian representation, joint 3DGS optimization, and cross-view alignment contribute to 3D-Consistent Instance Grounding.
Using the same form of cross-view supervision as IGGT, the frozen-3DGS variant achieves stronger track consistency but less accurate instance regions, suggesting that shared Gaussians promote cross-view consistency while frozen geometry leads to over-smoothed instance boundaries. 
Joint optimization improves both temporal consistency and rendering quality by allowing Gaussian geometry and opacity to adapt to instance supervision. 
Removing $\mathcal{L}_{\mathrm{cross}}$ also weakens instance tracking, confirming that explicit prototype alignment and the Gaussian representation jointly establish consistent instance identities. 
These results show that the Gaussian representation serves as an active 3D carrier for consistent instance grounding.

\paragraph{Semantic-Guided Instance Discrimination.}
Semantic-guided hard-negative reweighting is evaluated in~\cref{tab:multi_view_results,fig:IGGT_vis}.
Its gains under most input-view settings, together with the clearer separation of neighboring chairs, show that language-aligned semantic similarity focuses instance learning on confusing same-category objects. 
This demonstrates that semantic cues improve the discrimination of confusing same-category instances.

\paragraph{Instance-Grounded Semantic Aggregation.}
\Cref{tab:multi_view_results} ablates instance-grounded semantic aggregation. Enabling this module consistently improves mIoU and mAcc across input-view settings, while the qualitative results in~\cref{fig:uni3re1} exhibit more coherent object regions. These results show that instance regions provide stable object-level units for aggregating language-aligned semantics and promote object-level semantic consistency.

\paragraph{Boundary-Aware Reconstruction.}
The boundary-aware objective converts discontinuities in the learned instance features into spatially focused RGB supervision, directing optimization toward regions where smooth reconstruction decoders tend to blur adjacent objects. 
\Cref{tab:Ablation} shows that its effect on depth accuracy is stronger than on appearance quality, indicating that the improvement mainly comes from refining geometric discontinuities.
This result demonstrates the reverse interaction, in which instance structure improves 3DGS reconstruction near object boundaries.

Together, the four ablations provide consistent evidence for the effectiveness of 3D-Consistent Instance Grounding and the reciprocal interactions enabled by Instance-Centric Coupling.

\section{Conclusion}
We present \OurMethod, a unified feed-forward 3DGS framework that constructs an instance-aware Gaussian representation from pose-free multi-view images. 
The representation jointly encodes appearance, geometry, instance identity, and language-aligned semantics in shared Gaussians.
We further introduce a 3D-Consistent Instance Grounding module to learn renderable and cross-view-consistent instance features.
Building on the grounded instance structure, Instance-Centric Coupling enables reciprocal interactions between 3D reconstruction and instance-aware scene understanding, improving both reconstruction quality and scene-understanding performance. 
Comprehensive experiments on novel-view synthesis, instance segmentation, and open-vocabulary semantic understanding under varying input-view settings and on an unseen dataset demonstrate SOTA performance, practical efficiency, and strong generalization. 
A remaining limitation is that feature clustering becomes more expensive as the number of views and Gaussians increases. Future work will explore more compact scene representations.

\bibliography{main}
\end{document}